\documentclass[letter, 10pt, conference]{ieeeconf}
\IEEEoverridecommandlockouts

\usepackage{cite}

\usepackage{amsmath,amssymb,amsfonts}
\usepackage{algorithmic}
\usepackage{graphicx}
\usepackage{subcaption}
\usepackage{pdfpages}
\usepackage{textcomp}
\usepackage{xcolor}
\usepackage{bm}
\usepackage{booktabs}
\usepackage{svg}
\svgsetup{inkscapelatex=false}
\usepackage{tikz}
\usepackage{hyperref}
\def\BibTeX{{\rm B\kern-.05em{\sc i\kern-.025em b}\kern-.08em
		T\kern-.1667em\lower.7ex\hbox{E}\kern-.125emX}}
\usetikzlibrary{arrows,decorations.pathmorphing,backgrounds,fit,positioning,calc,shapes}

\usepackage[linesnumbered,ruled,algo2e,resetcount]{algorithm2e}%for the environment
\SetAlFnt{\small}
\SetAlCapFnt{\small}
\SetAlCapNameFnt{\small}

\begin{document}

% ------ DECIDE ON A TITLE!
\title{\LARGE \bf
A Framework for Learning Behavior Trees \\ in Collaborative Robotic Applications
\thanks{
\textbf{This work has been submitted to the IEEE for possible publication. Copyright may be transferred without notice, after which this version may no longer be accessible.}
This project is financially supported by the Swedish Foundation for Strategic Research and by the Wallenberg AI, Autonomous Systems, and Software Program (WASP) funded by the Knut and Alice Wallenberg Foundation. The authors gratefully acknowledge this support.}
}

%Authors metadata
\author{\authorblockN{
Matteo Iovino$^{a,b}$,
Jonathan Styrud$^{b,c}$,
Pietro Falco$^{a}$ and
Christian Smith$^{b}$}\\
\thanks{$^{a}$ABB Corporate Research, Västerås, Sweden}
\thanks{$^{b}$Division of Robotics, Perception and Learning, KTH - Royal Institute of Technology, Stockholm, Sweden}
\thanks{$^{c}$ABB Robotics, Västerås, Sweden}
}

\maketitle

\begin{abstract}
In modern industrial collaborative robotic applications, it is desirable to create robot programs automatically, intuitively, and time-efficiently. Moreover, robots need to be controlled by reactive policies to face the unpredictability of the environment they operate in.
In this paper we propose a framework that combines a method that learns Behavior Trees (BTs) from demonstration with a method that evolves them with Genetic Programming (GP) for collaborative robotic applications. 
The main contribution of this paper is to show that by combining the two learning methods we obtain a method that allows non-expert users to semi-automatically, time-efficiently, and interactively generate BTs.
We validate the framework with a series of manipulation experiments.
The BT is fully learnt in simulation and then transferred to a real collaborative robot.
\end{abstract}


\section{Introduction}
Modern industrial robots can solve complex tasks in controlled environments with high precision and reliability. However, trends in automation are pointing towards smaller production series with the robots program needing more frequent updates. At the same time, robots are increasingly operating in workspaces shared with humans, causing a more unpredictable environment. It is therefore desirable that new robot policies or programs can be created quickly without needing high programming skills and that the resulting programs are reactive to changes in the environment.
One way to alleviate the demand for programming skills is to have the user provide input in the form of demonstrations rather than code. The contributions of this paper is a framework that combines a method that evolves Behavior Trees (BTs) with Genetic Programming (GP) with a method that learns BTs from demonstration. By combining the two methods it is possible to learn BTs in an unsupervised fashion while exploiting human experience in task solving.

The method allows non-expert users to semi-automatically, time-efficiently, and interactively generate BTs for manipulation tasks. The interaction is facilitated by a user interface that allows users to start and stop the learning process at will as well as inputting demonstrations.

As a result, the shortcomings of the two individual components are resolved by their combination.

The code repository for this paper is available online\footnote{\url{https://github.com/matiov/BT-learning-framework}}.

%\begin{itemize}
%    \item a method to evolve BTs from user inputs in the form of demonstration of a robotic task
%    \item a method to increase the levels of fidelity of the simulator in order to improve the robustness of the learnt BTs
%    \item a comprehensive framework with a UI to allow the user to intuitively exploit the framework to teach a robotic task in simulation
%    \item examples showing that the framework can be utilize by non-expert users and that the learnt solution transfers to the real robot
%\end{itemize}

\begin{figure}[tbp]
    \centering
    \includegraphics[width=.6\linewidth]{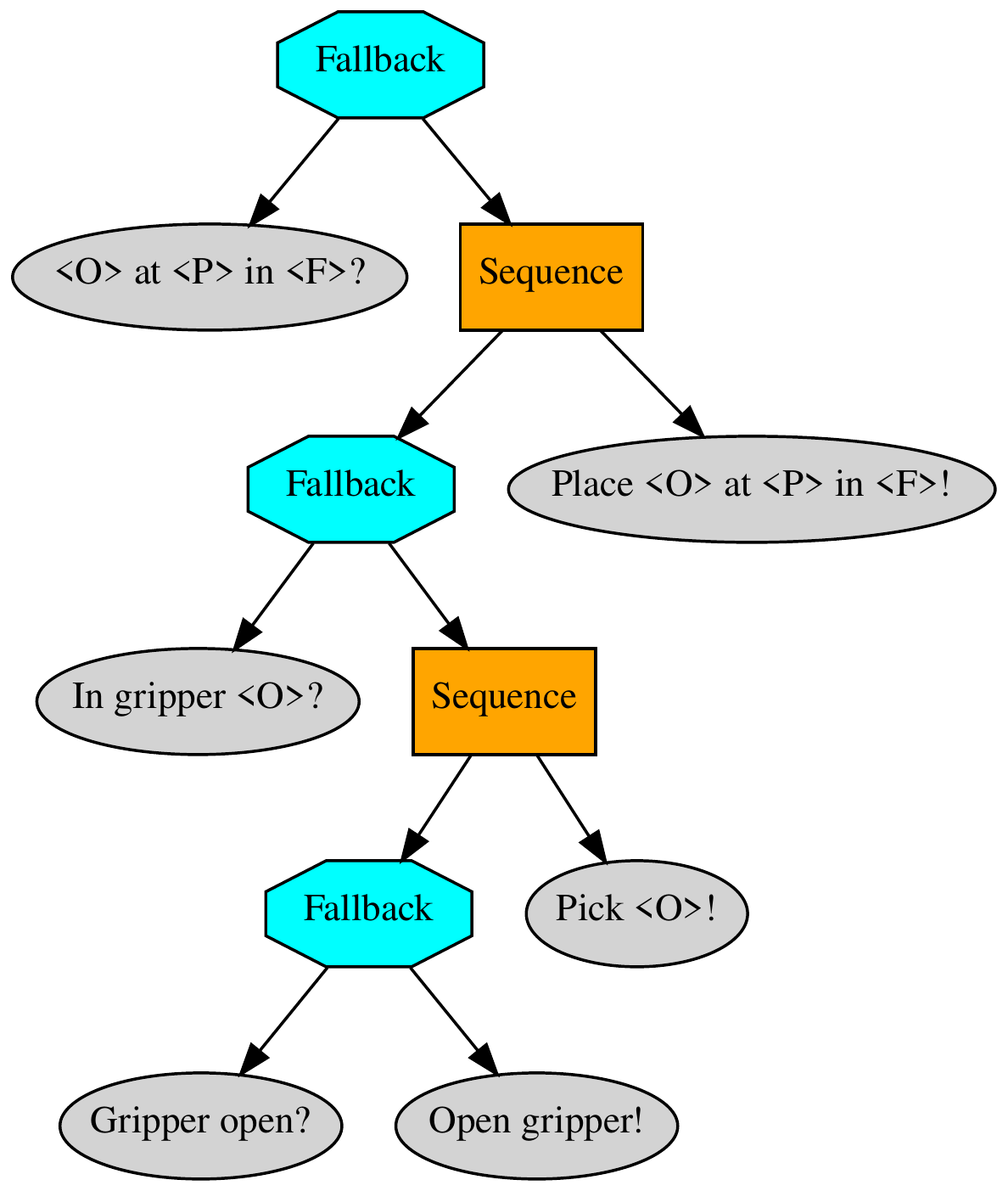}
    \caption{Subtree responsible for picking the item \texttt{<O>} and placing it at pose \texttt{<P>} in the reference frame \texttt{<F>}.}
    \label{fig:gene}
\end{figure}

\section{Background and Related Work}

\subsection{Behavior Trees}

BTs are a task switching policy representation originating in the gaming industry and later transferred to robotics~\cite{colledanchise_behavior_2018}.

A BT is represented as a directed tree where a tick signal originates from the root and propagates down the tree with a depth-first pre-order traversal. Nodes execute only when they receive the tick signal and return one of the status signals \textit{Success}, \textit{Failure}, and \textit{Running}. Internal nodes are called \emph{control flow nodes} (polygons in Figure~\ref{fig:gene}). The most common types being \emph{Sequence}: runs children in a sequence, returning once all succeed or one fails, and \emph{Fallback} (or \emph{Selector}): runs children in a sequence, returning when one succeeds or all fail. Leaves are called \emph{execution nodes} or \emph{behaviors} (ovals in Figure~\ref{fig:gene}) and are of type Action(!) or Condition(?). The former encode robot skills while the latter encode status checks and sensory readings, thus immediately returning \textit{Success} or \textit{Failure}.

BTs have explicit support for task hierarchy, action sequencing, and reactivity~\cite{iovino_survey_2022}. They are modular by design because every element shares the same infrastructure: every node receives the tick as input and outputs the return statuses. Moreover, modularity allows every building block to be independently tested and reused. The \emph{Running} return state grants the reactivity property because a running action can be preempted by higher priority ones.

BTs improve on other task plans representations, such as Finite State Machine, especially in terms of modularity and reactivity~\cite{iovino_programming_2022,colledanchise_how_2017,biggar_modularity_2022}.

\begin{figure}[tbp]
    \centering
    \includegraphics[width=\linewidth]{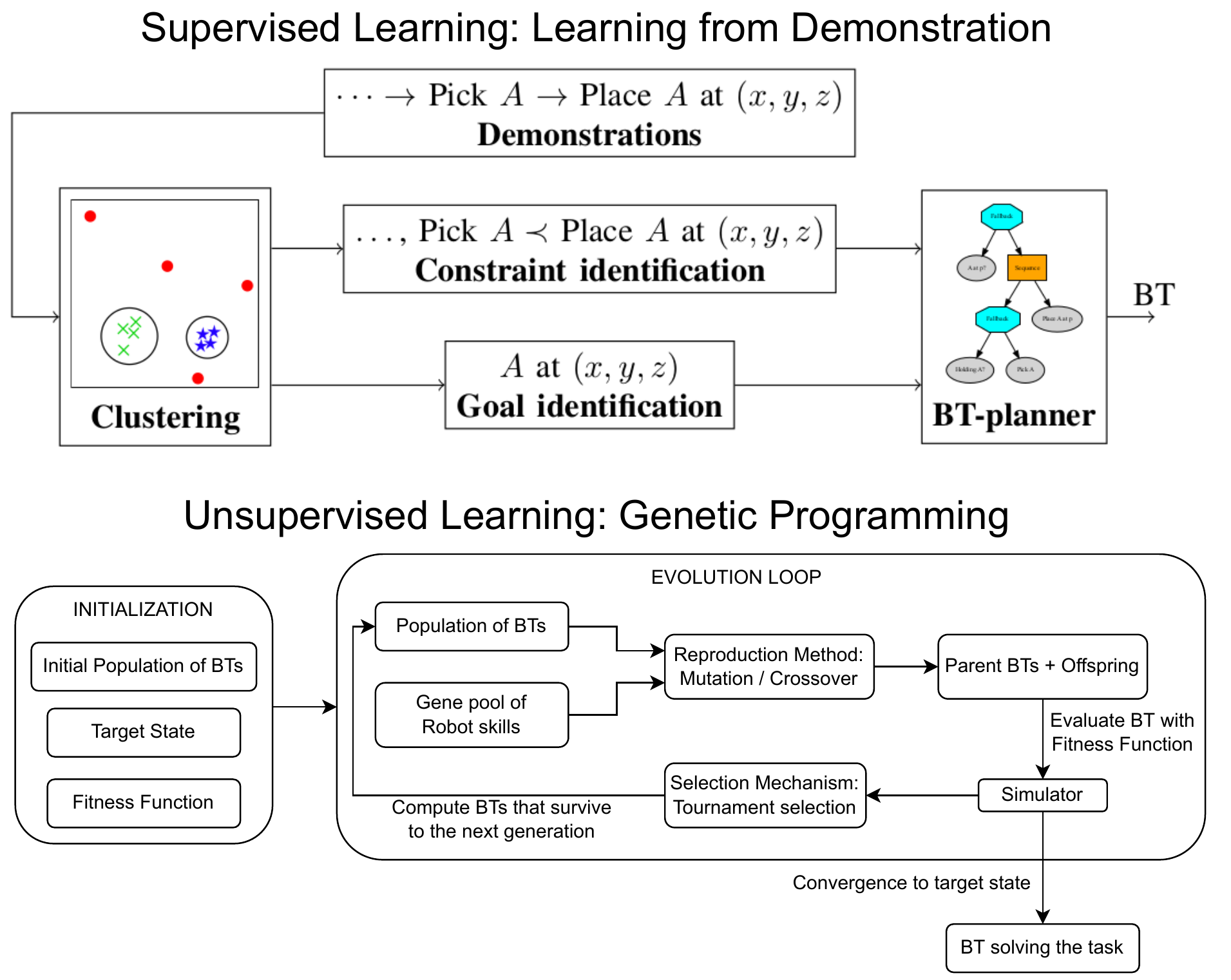}
    \caption{Functional scheme of the proposed learning methods for intuitive and efficient robot behavior creation for high-level control~\cite{iovino_behavior_2022}.}
    \label{fig:learning_BT}
\end{figure}

\subsection{Evolving BTs with Genetic Programming}

Genetic Programming (GP) is an unsupervised optimization algorithm that can evolve programs represented as trees~\cite{langdon_genetic_1970} using the operators \emph{crossover} and \emph{mutation}. Individuals are evaluated by performing a task in the environment they operate in through a fitness function and selection methods that decide which individuals survive and produce offspring. GP is particularly fit to learn BTs because their modularity facilitates reproduction operations, they are a tree representation, and they can be serialized to reduce the computational effort of the evolutionary process.

The GP algorithm illustrated in Figure~\ref{fig:learning_BT}-bottom is based on our prior works~\cite{iovino_learning_2021, styrud_combining_2022} and inspired by~\cite{koza_genetic_1994}. The main evolutionary steps are:
\begin{enumerate}
    \item \textbf{Reproduction:} When generating the offspring, the mutation operation can add, delete, or change nodes with a user selected probability to choose a control or execution node. With crossover a random subtree is selected from each of two parent individuals and inserted in the other individual at a random point. The modularity of BTs is the key feature of this step because subtrees can be moved without compromising the logical functioning of the tree.
    \item \textbf{Evaluation:} Each individual is run in the simulation environment and a fitness function is evaluated to assess the individual's score. The design of the fitness function has to trade-off generalizability with specificity. A fitness function that is too task specific can improve the convergence rate but it is not re-usable, thus increasing the engineering effort. Similarly to~\cite{styrud_combining_2022}, we consider the Euclidean distance between objects current and goal positions, the size of the tree, the number of ticks required to run the individual, and if the BT terminates with a \textit{Success} or \textit{Failure} state.
    \item \textbf{Selection:} To determine the individuals that survive to the next generation we perform tournament selection, where individuals are compared in randomly assigned pairs. Out of each pair, only the best scoring individual survives. This method grants high recombination of genes but it always keeps the best scoring individual and discards the worst one. 
\end{enumerate}

This approach is explainable in the sense that each step of the algorithm is controllable and predictable. Moreover, it is easy to implement. The main shortcoming is the high number of learning episodes required to converge. Therefore, it benefits from a fast simulator since every individual must be tested and evaluated. In~\cite{styrud_combining_2022}, we mitigated this problem by bootstrapping the GP with planned BTs solving subtasks.

Our approach uses similar constraints as in~\cite{mcclarron_effect_2016}. We avoid to use the same type of control node on two consecutive levels of the tree, conditions on the rightmost position of a sub-tree, child-less control nodes, or having identical nodes next to each other, as all these are unnecessary variations of the BT.
Unlike~\cite{colledanchise_learning_2019}, we do not constrain mutation to nodes of the same type, to increase diversity. We also let the GP algorithm find the conditions to use instead of explicitly specifying them. Some conditions that actions must always check are included within the action behaviors.

\subsection{Learning BTs from Demonstration}

Learning from Demonstration (LfD) methods allow robots to learn programs from human demonstrations~\cite{ravichandar_recent_2020}. This method is especially useful when users do not have enough programming skills, or writing robot programs to solve a task takes too long. Demonstrations can be given in the form of \textit{kinesthetic teaching}, where the user physically moves the robot, \textit{teleoperation}, where a robot is controlled through an external device\----particularly useful when the robot operates in unreachable environments\----and \textit{passive observation}, where the robot or the human are endowed with tracking systems and the demonstrator's body motion is recorded. These methods are often intuitive and little to no training is required for the user. 

Our method to learn BTs from demonstration proposed in~\cite{gustavsson_combining_2022} and schematically outlined in Figure~\ref{fig:learning_BT}-top, builds on the following steps:
\begin{enumerate}
    \item The user inputs the demonstrations through the user interface as a sequence of actions that brings items in the environment from a starting to a goal configuration. For every performed action in the demonstration, we record the pose of the end-effector in the reference frame of all the relevant objects for the task as well as the type of the action performed.
    \item Similar actions across different demonstrations of the same task are clustered together to infer their frame of reference. The clustering algorithm requires at least 3 demonstrations to infer the reference frame, otherwise the base frame of the robot is used as default value.
    \item It is assumed that the user solves the task by the end of the demonstration. Therefore, the configuration of the items when the last action is performed is used as goal condition. Other task constraints on the order of the actions are inferred as well.
    \item A planner similar to the one defined in~\cite{colledanchise_towards_2019} is used to automatically generate the BT with backchain design, provided that the actions are defined together with their pre- and post-conditions. With backchaining, starting from the goal, pre-conditions are iteratively expanded with actions that achieve them\----those actions that have that particular condition as their post-conditions. Then, those actions' unmet pre-conditions are expanded in the same way.
\end{enumerate}

In~\cite{iovino_interactive_2022} we extend this method to allow the robot to ask the user follow-up clarification questions if there are ambiguities in the task to solve. For instance, when there are equivalent items available for the robot to pick. 

The main shortcoming of this method is that it does not generalize well when the task to demonstrate requires many actions. Performing several demonstrations might be tedious and source of errors, especially when combined with perception algorithms whose performance is affected by lighting conditions and object occlusion.

\subsection{Related Work}
%\todo{Shorten related works if we run out of space}

Evolutionary approaches to learn BTs were first applied to computer games~\cite{hutchison_evolving_2010, colledanchise_learning_2019, perez_evolving_2011, nicolau_evolutionary_2017, zhang_behavior_2018, zhang_learning_2018, mcclarron_effect_2016} but later years have also seen it applied to robotic applications~\cite{scheper_behavior_2015, jones_evolving_2018, jones_two_2018, neupane_learning_2019,iovino_learning_2021, styrud_combining_2022}.

\par
Authors in~\cite{perez_evolving_2011, nicolau_evolutionary_2017} successfully use Grammatical Evolution (GE). With GE, the design of the grammar requires domain knowledge and the engineering effort grows with the complexity of the task. Further, the disconnect between genotype and phenotype can make analysis and implementing heuristics difficult and hinders locality~\cite{rothlauf_locality_2006, hallawa_evolving_2020}.
In~\cite{zhang_behavior_2018, zhang_learning_2018, mcclarron_effect_2016}, structural and dynamic constraints were implemented in the GP, speeding up the learning by preventing the generation of undesirable trees. Similar constraints are also used in our implementation.
In~\cite{colledanchise_learning_2019}, GP was combined with a local search but this is harder to re-implement in a non-deterministic robotic scenario. Most notably,~\cite{styrud_combining_2022} presented a method to combine the results from a planner with a GP algorithm. An adaptation of that method is what we use in the proposed approach of this paper to seed the GP with the information gathered in the demonstrations.

There has been some previous work on combining LfD and BTs~\cite{sagredo-olivenza_trained_2019, french_learning_2019, robertson_building_2015, suddrey_learning_2022, gustavsson_combining_2022, iovino_interactive_2022}. The method proposed in~\cite{sagredo-olivenza_trained_2019, french_learning_2019}, and~\cite{ wathieu_rebt-espresso_2022} learns a Decision Tree (DT) to map from state space to action space. The DT is then converted into a BT using the fact that BTs generalize DTs~\cite{colledanchise_how_2017}. In~\cite{robertson_building_2015}, the method directly encodes the demonstrated sequence of actions as a BT and in~\cite{suddrey_learning_2022} BTs are generated by natural language instructions.

In~\cite{sagredo-olivenza_trained_2019}, LfD was used to assist in creating behaviors for Non-Player Characters (NPCs) in computer games. A DT was trained as a policy choosing the NPCs next action depending on the game state. The DT was flattened into a set of rules, then simplified and translated into a BT. The BT required a final tuning of its parameters, thus limiting its usage. This was extended by~\cite{french_learning_2019} and improved in~\cite{wathieu_rebt-espresso_2022} who generalized it to use any logic minimizer on the DT. The solution is implemented on a mobile manipulator to perform a house-cleaning task. In this work, the whole action space and state space were encoded in the final tree which would then contain elements that are not used at run-time and unnecessarily complicate the structure. Further, the frame of reference was fixed in the actions which limits their reusability. The method does not make use of any previous knowledge of the behaviors and it can only execute behaviors that were demonstrated. 

In~\cite{robertson_building_2015} the BT was synthesized directly. An agent was trained to play the video game StarCraft from expert demonstrations. Each demonstration resulted in a sequence of actions placed under a sequence node in the BT, then all sequence nodes for each demonstration were placed under a fallback node. Finally, the BT was simplified by finding similarities between different demonstrations. This approach results in large and hard to read BTs ($>50.000$ nodes) and limits the reactivity as in-game actions and conditions are all put under the same Sequence node.

Finally, in~\cite{suddrey_learning_2022} authors proposed a method for generating BTs from natural language instructions. The method parses the instruction and searches a database for trees solving the requested task. If none are found, a new tree is learned by matching the parsed expression to hand-coded primitive methods as simple BTs. Although the method allows to define a wide variety of tasks, the learned trees are hard to read and it is not clear how the method would handle tasks where the relative position of objects in the scene matters.

Other LfD methods learn task plans without using BTs, but instead using a Finite State Machine~\cite{niekum_learning_2015} or a Hidden Markov Model~\cite{hovland_skill_1996}. In~\cite{konidaris_robot_2012} plans were created by chaining skills that have an effect that allows the next skill to be executed successfully. Chains from multiple demonstrations were combined into a skill tree with multiple chains that achieve the same goal. The main advantage of using BTs over these methods is the inherent reactivity and increased readability of BTs.

The method in this paper mainly builds upon~\cite{gustavsson_combining_2022} which in turn was built on the work of\cite{ekvall_robot_2008} and~\cite{colledanchise_towards_2019}. The method presented in~\cite{ekvall_robot_2008} uses demonstrations to gather information on order of actions. With multiple demonstrations it can also learn to generalize instead of just repeating the demonstrations. However, the built plan is fixed and not reactive to external disturbances.

The later work of~\cite{colledanchise_towards_2019} presented a planner to iteratively build a BT. In each iteration the current tree is executed and failing pre-conditions are replaced with subtrees that execute actions with appropriate post-conditions, a method called backchaining. The BT is built at run-time and the result is therefore not deterministic and depends on the state of the environment. After that,~\cite{gustavsson_combining_2022} expanded on the backchaining idea while also incorporating LfD and learnt also reference frames and context of each behavior using a clustering approach, thereby making more types of tasks solvable. However, it still relies on having a planner that is able to solve the task at hand.

\section{Proposed Method} \label{sec:method}

%\begin{itemize}
%    \item GP settings and parameters
%    \item LfD settings (how demonstrations are provided in the simulator)
%    \item Simulator settings (different LVs of fidelity)
%\end{itemize}

\begin{figure}[ht]
    \centering
    \includegraphics[width=\linewidth]{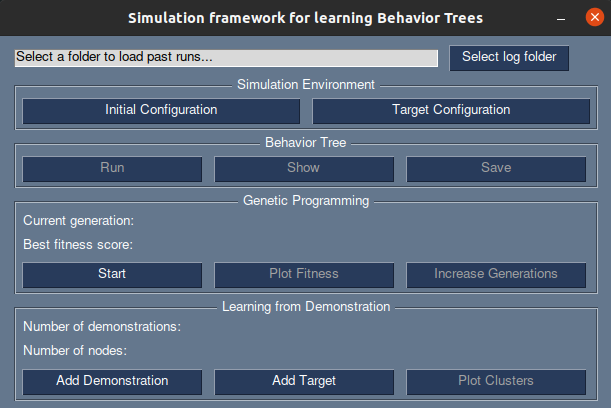}
    \caption{Graphical User Interface to control the GP algorithm, the LfD framework, and to run and display the learnt BT.}
    \label{fig:gui}
\end{figure}

The method of this paper combines BTs learnt from user demonstrations with ones evolved by GP in an interactive way.
On a higher level, our method alternates demonstration steps with evolutionary ones. The idea is that user demonstrations can be exploited to bootstrap the GP algorithm, to edit the output from the GP, or to provide new information to the GP if the evolutionary process is stuck in local optima. For this reasons, we let users to interact with the learning framework through the user interface in Figure~\ref{fig:gui}, where they can input demonstrations, start-stop-resume the GP, and visualize or run the learnt BT. We therefore have a framework that automatically generates BTs while exploiting user experience in task solving.

\begin{figure}[ht]
    \centering
    \includegraphics[width=\linewidth]{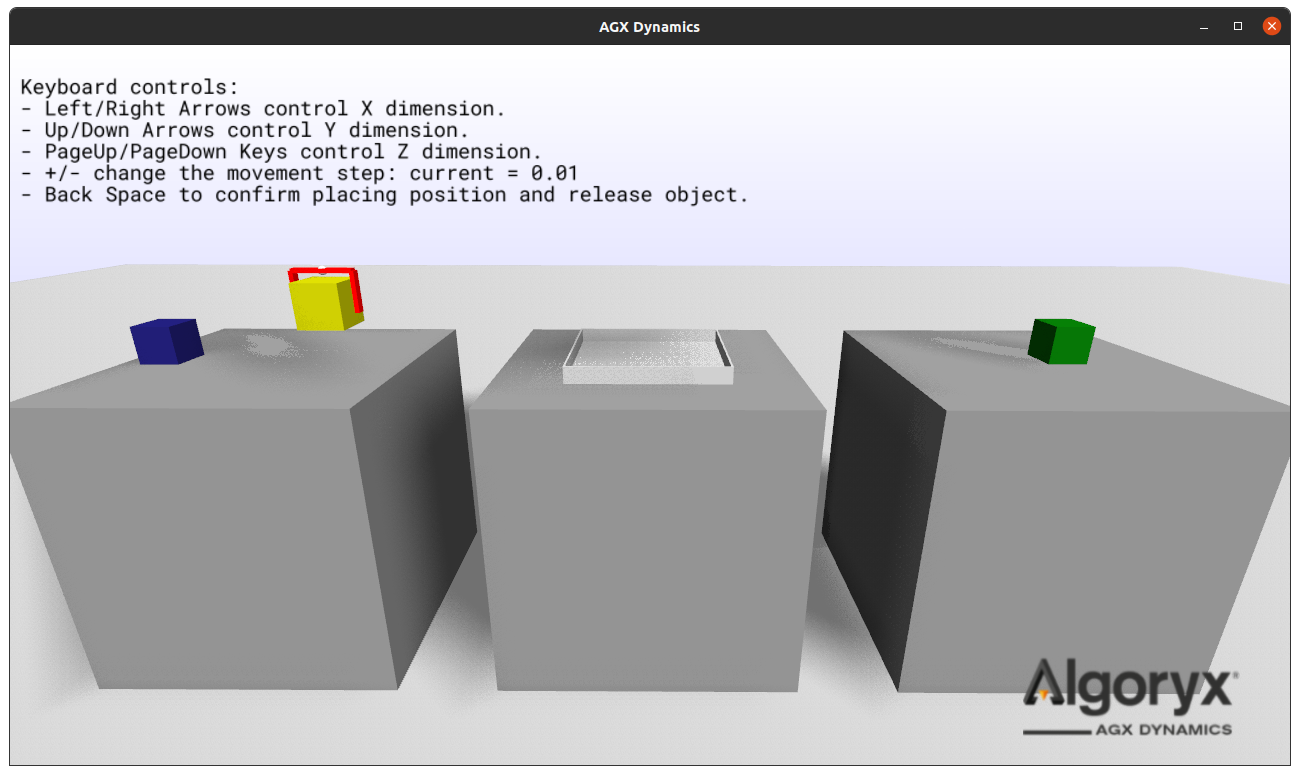}
    \caption{Demonstration view of the simulator. The user can control the gripper with the keyboard and place the items at the desired position.}
    \label{fig:demo}
\end{figure}

%More in detail, unlike~\cite{gustavsson_combining_2022} and~\cite{iovino_interactive_2022}, 
The demonstrations are provided in simulation, by selecting the type of action to perform:
\begin{itemize}
    \item \textbf{Pick:} a drop-down menu is generated allowing the user to select the item to pick. This action will make the gripper to grasp the selected object;
    \item \textbf{Place:} the user takes control of the gripper that can be teleoperated through the keyboard and moved to the desired position, as shown in Figure~\ref{fig:demo}. Then, the gripper releases the grasped object.
\end{itemize}

A BT generated this way to pick and place an item is shown in Figure~\ref{fig:gene}.

For the gene pool of the GP algorithm, we combine pick and place of the same object in a single subtree that we consider as a behavior/gene (Figure~\ref{fig:gene}). This is due to the fact that for every movable item (e.g. the boxes) we sample 5 target positions in the reference frame of every other item (corresponding to left, right, above, below, and on top), plus 9 target positions inside the kit box. The reason behind this design choice is to constrain the search space, and we can assume that such a list of hand-coded/learnt subtrees for standardized actions are provided beforehand. Nonetheless, the gene pool counts 61 behaviors (56 for pick and place objects plus 5 related to the gripper). 

For the design of the fitness function, we follow the guidelines of~\cite{styrud_combining_2022}. Since the tasks consist of moving objects,  we base the fitness function on the Euclidean distance between objects current and target positions. A factor $\lambda$ penalizes larger trees over smaller ones thus preventing the learnt solution from bloating. Then, $\tau$ is removed from any tree that ends by timeout to prevent the GP from exploiting the fact that we interrupt the simulation of an individual after a certain amount of time. Finally, $\phi$ is removed from trees that end in Failure state. Similarly to~\cite{styrud_combining_2022}, the fitness function can be thus modelled as
\begin{equation}
    \mathcal{F} = - \sum_\mathcal{O} max(0, ||o - g||) - \lambda L - \tau T - \phi F
\label{eq:fitness}
\end{equation}
where $\mathcal{O}$ is the set of objects with current positions $o$ and goals $g$, with $||o - g||$ the distance error in mm. $L$ is the number of nodes. $T$ is 1 if the tree ended by timeout and 0 otherwise and $F$ is 1 if the tree ended in Failure state and 0 otherwise. The values used for the experiments of Section~\ref{sec:exps} are reported in Table~\ref{tab:gp_pars}. Note that the goal of the paper is to show the benefits of the combined GP and LfD approaches. By combining the two approaches, less effort is required in finding an optimal choice for the parameters through hyperparameter search.

The simulation environment builds on AGX Dynamics from Algoryx\footnote{\url{https://www.algoryx.se/agx-dynamics/}}.
In prior work~\cite{iovino_learning_2021} we designed a probabilistic state machine as simulator to make the simulation faster and the problem tractable. With AGX Dynamics instead, we can simplify the simulation environment at will. For the experiments, we idealize the robot by simulating only a gripper that can teleport directly above the target. In this way, we simulate only the physical properties that are relevant for the task, such as contacts between objects and gravity, disregarding the rest. We achieve a simulation rate that is 100 times faster than real-time when the graphics are disabled. Having the possibility to decide the levels of fidelity, allows users to test if the learnt solution also works in a model of the environment closer to reality. In this paper, we skip this step by transferring the solution directly to the real robot.

From what concerns the implementation, the GUI controls the simulation through processes. Therefore, it is possible to run several instances of the simulation in parallel.

To conclude, by combining the two methods we address the shortcomings mentioned in the previous section.
\begin{enumerate}
    \item We obtain an interactive framework rather than a monolithic one that applies the different methods in series, as we did in~\cite{styrud_combining_2022}.
    \item We are able to stop the GP algorithm at will, if for instance the fitness score doesn't improve, and input more useful data in the form of demonstrations.
    \item We are able to demonstrate small subtasks and then let GP algorithm solve the full problem, thus improving the generalizability of the LfD method.
    \item We are able to modify the target for the GP algorithm by inputting new demonstrations.
    \item By teaching in simulation, we do not need to have access to the physical robot or to rely on the perfect functioning of a perception algorithm.
\end{enumerate}

\section{Experiments and Results} \label{sec:exps}

%\begin{itemize}
    %\item proof of concept: perform experiments with the 3 tables, random initial condition %and different lvs of simulator fidelity (stacking 3 boxes in the kit box?)
    %\item perform experiments for several tasks (pick examples from the LfD paper so that we can compare BTs from that paper to the ones in this and see if the GP brings improvements in terms of robustness)
    %\item use the simulation of the WARA lab to perform a kitting task so that we can do sim-to-real transfer
%\end{itemize}

\begin{table}[htbp]
\tiny
\scriptsize
\caption{GP parameters}
\begin{center}
\begin{tabular}{|c|c|}
\hline
\bf{Parameter description} & {\bf Value} \cr
 
\hline
Population size & 16 \cr
\hline
%//Initial random BT size & 8 \cr
%\hline
Parents selected for mutation & 12 \cr
\hline
Number of mutation offspring per parent & 2 \cr
\hline
Maximum number of mutations per individual & 3 \cr
\hline
Mutation probabilities for add, delete, change & 10\%, 50\% and 40\%\cr
\hline
Parents selected for crossover & 4 \cr
\hline
Number of crossover offspring per parent & 2 \cr
\hline
Selection method (reproduction and survival) & Tournament selection \cr
\hline
Number of elites & 2 \cr
\hline
Length penalty for each node ($\lambda$) & 10 \cr
\hline
Penalty for ending by timeout ($\tau$) & 30 \cr
\hline
Penalty for ending in Failure state ($\phi$) & 50 \cr
\hline
\end{tabular}
\end{center}
\label{tab:gp_pars}
\end{table}

In this section we support the claims made in Section~\ref{sec:method} and we consider a range of simulated manipulation tasks as example for a proof of concept. 

The manipulation tasks are examples of kitting tasks, where the robot has to place several items in a kit box. Kitting tasks are commonly performed by mobile manipulators where the robot navigates to collect items from shelves to place in a kit box. Here we focus on the manipulation part, and leave the navigation part for future work.

We perform three different experiments to highlight the strengths of our method. The workflow of the experiments is the following:
\begin{enumerate}
    \item the user let the GP algorithm run until it converges to output a BT that solves a user-defined task;
    \item the user adds another target or modifies the current one by inputting a demonstration;
    \item the user runs the GP algorithm again but using also the BT learnt from demonstration as baseline, similarly to what what was done in earlier work~\cite{styrud_combining_2022}.
\end{enumerate}

For Experiments~1 and~2 the GP is stopped after 50 generations and run for another 50 after inputting the demonstration.
With the simulator described in the previous section and the parameters for the GP algorithm of Table~\ref{tab:gp_pars}, running 50 generations takes approximately 15 minutes.
These parameters have been chosen based on our previous works on GP~\cite{iovino_learning_2021, styrud_combining_2022}.
Finally, to validate our method we transfer the solution learnt in simulation to a real robot. For the experiments we use an ABB YuMi robot.

\begin{figure}[ht] 
    \begin{center}
    \begin{subfigure}[t]{.40\linewidth}
    \includegraphics[width=\linewidth]{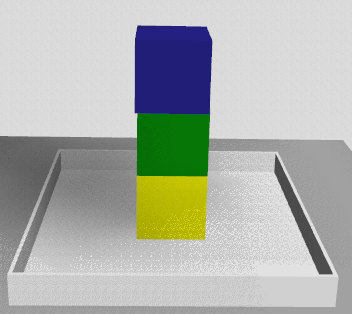}
    \caption{Experiments 1, 3, and 4.}
    \label{fig:exp1}
    \end{subfigure} \qquad
    \begin{subfigure}[t]{.48\linewidth}
    \includegraphics[width=\linewidth]{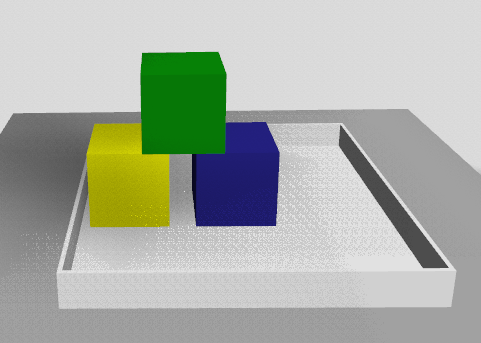}
    \caption{Experiment 2.}
    \label{fig:exp2}
    \end{subfigure}
    \caption{Target configurations for the experiments.}
    \label{fig:experiments}
    \end{center}
\end{figure}

\begin{figure*}[ht!]
\centering
\begin{subfigure}{0.32\textwidth}
\includegraphics[width=\textwidth]{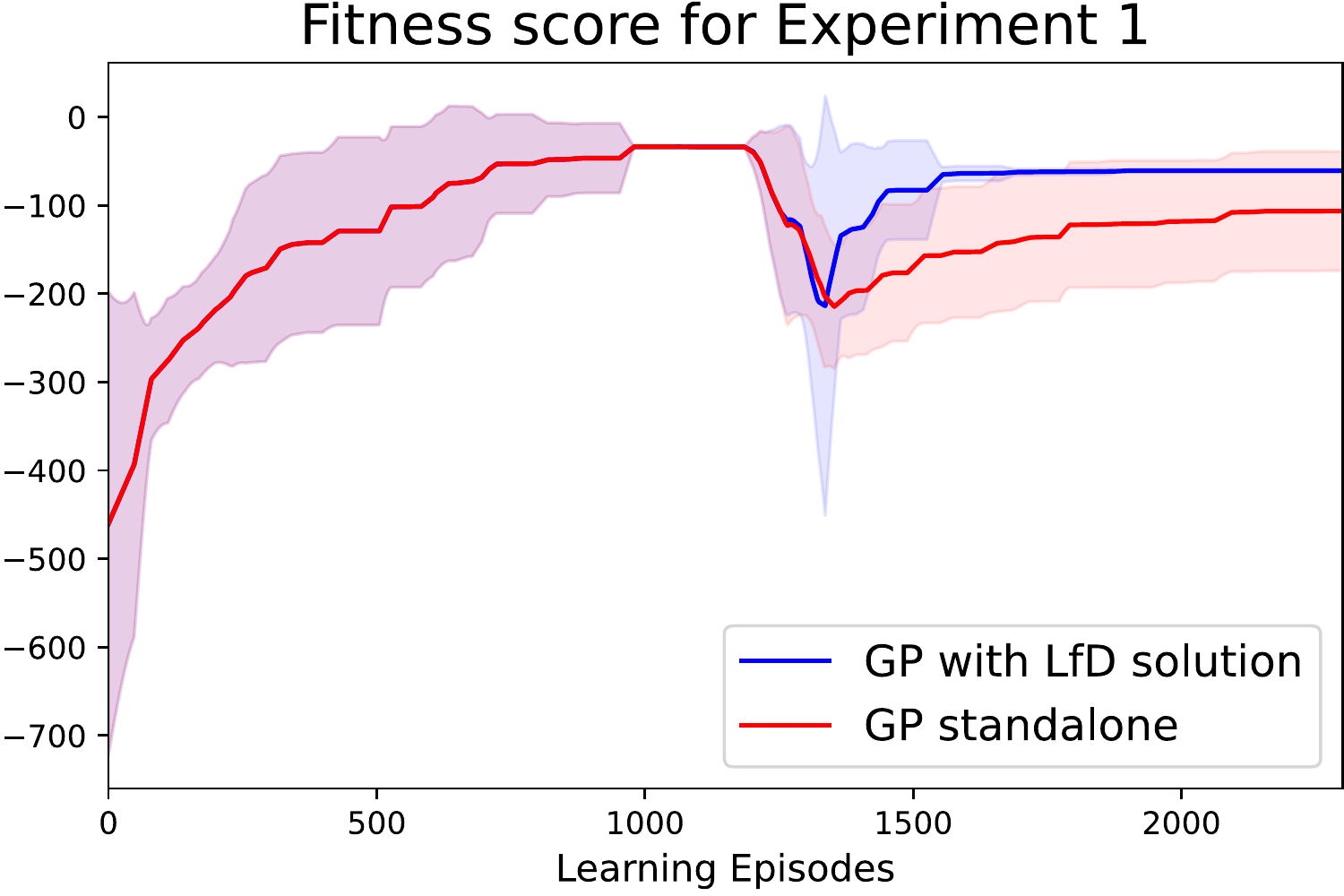}
\caption{Experiment~1: stacking boxes in the \textit{KittingBox} in a specific order.}
\label{fig:res1}
\end{subfigure}
\begin{subfigure}{0.32\textwidth}
\includegraphics[width=\textwidth]{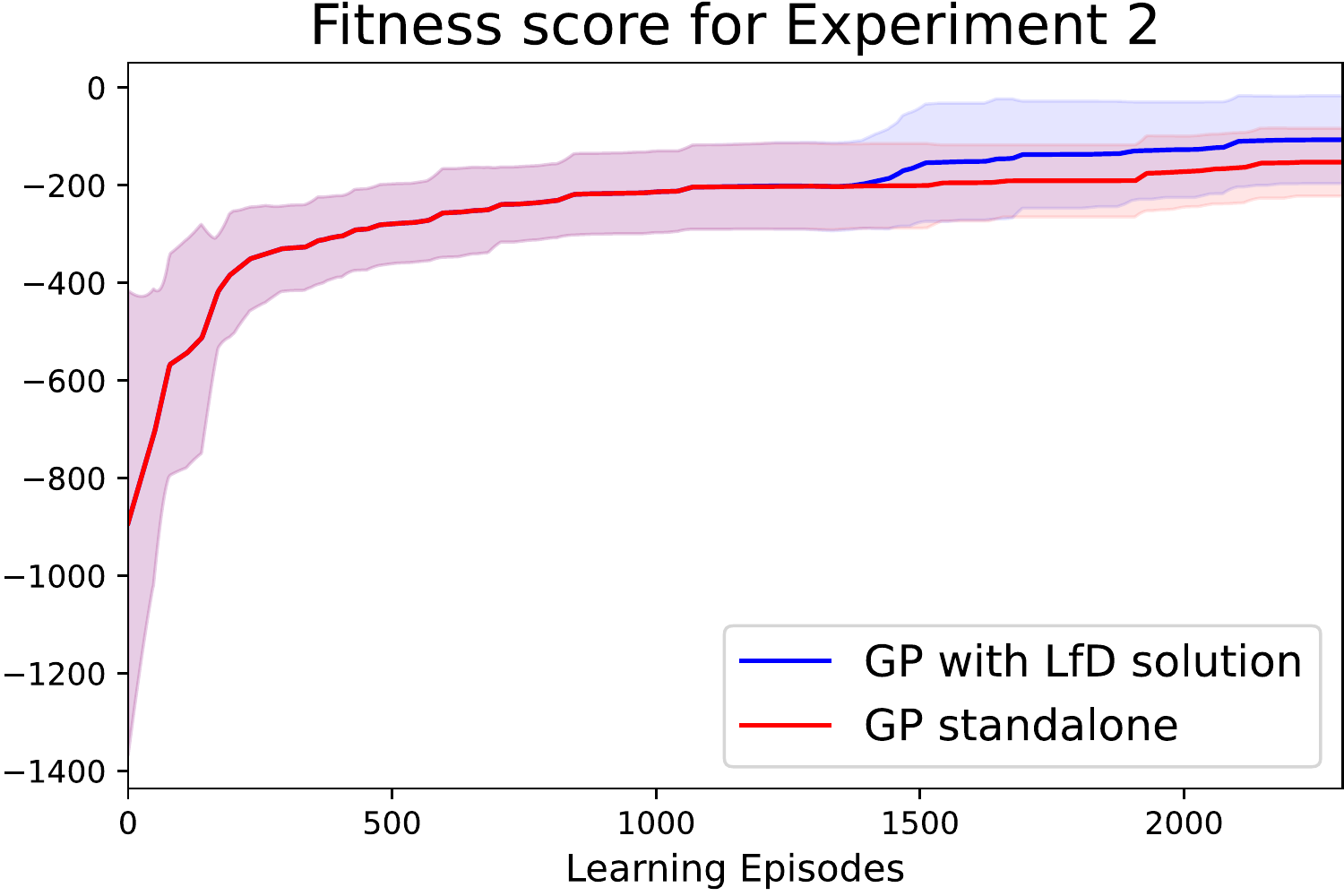}
\caption{Experiment~2: placing boxes in a pyramid configuration in the \textit{KittingBox}.}
\label{fig:res2}
\end{subfigure}
\begin{subfigure}{0.32\textwidth}
\includegraphics[width=\textwidth]{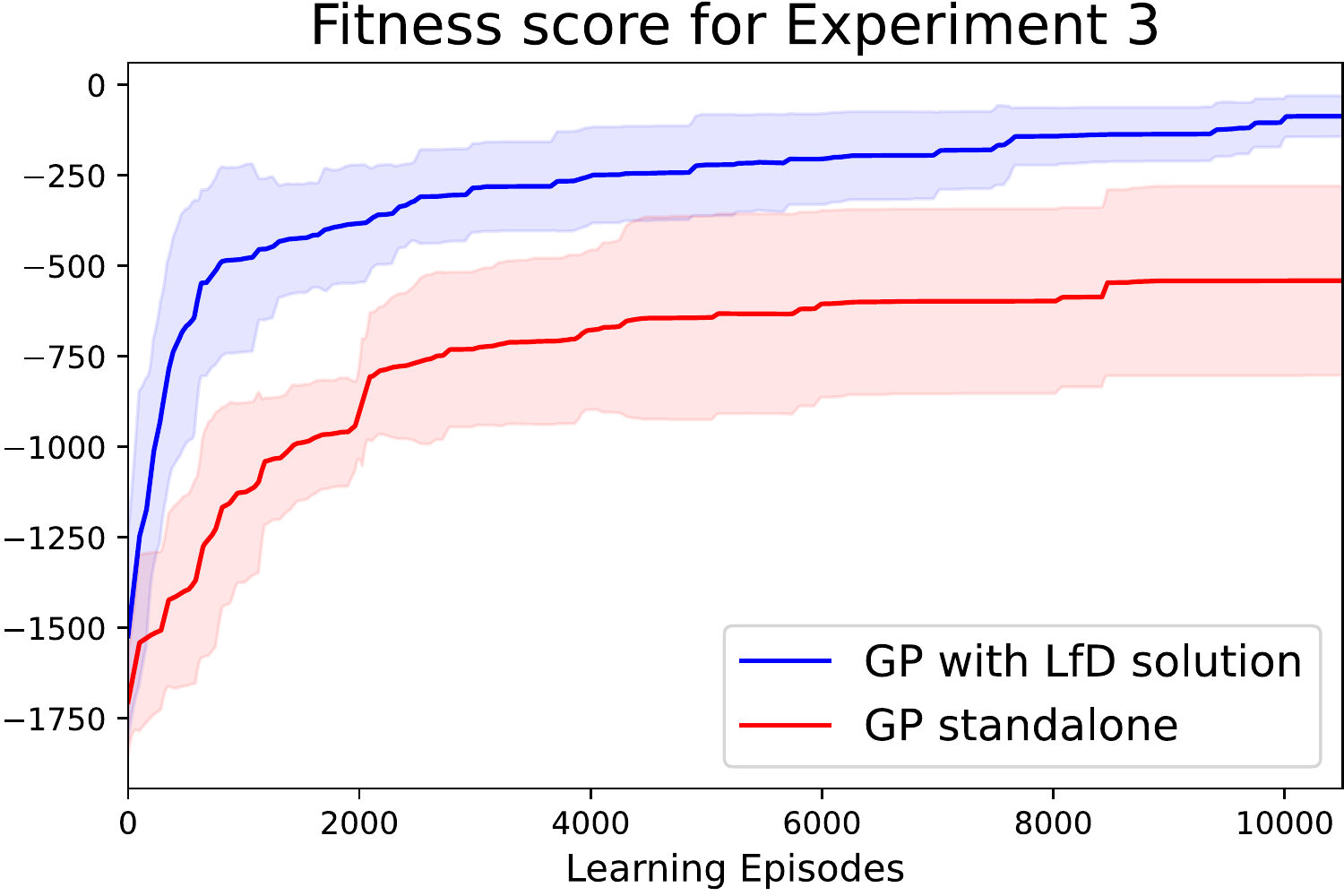}
\caption{Experiment~3: restacking boxes in the \textit{KittingBox} in a specific order from a wrong stack configuration.}
\label{fig:res3}
\end{subfigure}
\caption{Learning curves for experiments~1 to~3 averaged on 10 runs.}
\label{fig:results}
\end{figure*}

\paragraph{\textit{Experiment 1}} \label{exp1}

The goal of this experiment is to show that by inputting a demonstration, the user is able to modify the target of the task that the GP algorithm has previously solved. Neither the solution found by the GP nor the demonstration alone are sufficient to solve the modified task but only their combination. The GP has output a BT solving a different target while the demonstration shows only a limited part of the full task.

The target of the manipulation task is to stack two boxes in the kitting box. The robot has to place the \textit{YellowBox} in the center of the \textit{KittingBox} and the \textit{GreenBox} on top of the \textit{YellowBox}. The user then adds as a new target that the \textit{BlueBox} is placed on top of the \textit{GreenBox}, as shown in Figure~\ref{fig:exp1}. The GP runs for 50 generations to solve the first part of the task. After the target is changed it runs for 50 generations more.

\textbf{Results:} The learnt BT for this task features a Sequence node as a root and three children subtrees like the one in Figure~\ref{fig:gene}, with the appropriate values for the parameters \texttt{<O>}, \texttt{<P>}, and \texttt{<F>}. In Figure~\ref{fig:res1} we show the learning curve for the Experiment~1. After 50 generations ($\simeq 1300$ learning episodes) the fitness score drops because we introduce a new target and the previously learnt BT is no more a good solution.
In blue, the GP is boosted by a BT learnt from demonstration that performs the final step to bring the items from the old final configuration to the new one. In this case, the BT learnt from demonstration moves the \textit{BlueBox} on top of the \textit{GreenBox}. In red, we let the GP solve the task with updated target, without inputting any new information. The GP converges to a lower fitness score than before changing the target because it learns a larger tree.
This experiment supports the findings of~\cite{styrud_combining_2022} because it shows that the GP algorithm converges faster when boosted by a solution that solves parts of the task. Therefore, the GP algorithm benefits from solutions learnt from demonstration. Moreover, we show that the LfD method benefits from the evolutionary process because it relieves the user from demonstrating the whole task, but only the final part. 

\paragraph{\textit{Experiment 2}} \label{exp2}

In this experiment we show that a user can exploit demonstrations to achieve target configurations for the items that are not allowed by the set of behaviors available to the GP algorithm. In this case, we want to place two boxes side-by-side and place the third one on top of them in the middle, as shown in Figure~\ref{fig:exp2}. 

\textbf{Results:} The behavior for placing the third box in the specific target configuration of Figure~\ref{fig:exp2} is not defined in the initial set of the GP behaviors. Therefore, the GP algorithm can only solve the task sub-optimally. The two sub-optimal solutions are stacking the \textit{GreenBox} on top of either the yellow one or the blue one. If the user inputs demonstrations to solve the task correctly, then the learnt BT is inserted as baseline for the evolution and thus accessible to the GP for crossover and mutation. In this way, the GP can improve on the previous solution, as shown in Figure~\ref{fig:res2}. Since the difference between the correct solution and the sub-optimal ones is roughly $0.05~cm$ (half the size of the boxes) on the pose of the \textit{GreenBox}, the improvement of the fitness score that we observe in Figure~\ref{fig:res2} is not substantial.

\paragraph{\textit{Experiment 3}} \label{exp3}

The goal of this experiment is to show that a user can decide to stop the execution of the GP algorithm when it gets stuck in local optima to then help it converging by inputting information in the form of a demonstration. The goal of this task is the one in Experiment~1 (Figure~\ref{fig:exp1}) but instead of starting from initial random positions, the boxes are stacked in one of the tables with the wrong goal configuration. Therefore, the robot has to unstack them first and then stack them in the correct order in the \textit{KittingBox}. If given enough time, the GP will eventually solve the task because in every generation there is a non-zero probability that it finds the solution. However, especially in an industrial scenario the generation of robot programs is time constrained.

We let the GP run for twice as many generations and twice as large population as for the previous experiments (100 generations and 32 individuals in the population, respectively). We also modify the mutation probabilities to delete and change to $30\%$ and $60\%$ respectively to allow the GP to explore more widely the search space.

\textbf{Results:} The GP converges to a local optima where it places the boxes in the \textit{KittingBox} but not in the correct target configuration. Therefore, we stop the evolutionary process after 100 generations.

At this point, we perform a demonstration to show the robot how to unstack the boxes. In this case, we perform only one demonstration because it doesn't matter where the boxes are placed while unstacking. Therefore, it is enough if the robot learns to place them in its reference frame, which is the default option for the clustering step in the LfD method.
We also modify the fitness function, by adding a reward for unstacking the boxes successfully. This is done by capturing the goal of the subtask the demonstration is solving and assigning a score for this intermediate configuration of the items. In this paper we do this manually, but it could easily be automated. In this way we reward the GP for using the demonstrated BT correctly.
Moreover, we allow the user to decide where to insert the demonstrated BT during crossover. Since unstacking the boxes is the first step in the solving process, the demonstrated BT is inserted as first children of the root.
Finally, we run the GP for another 200 generations with the demonstrated BT as baseline, but we do not use the information from the prior runs. This is motivated by the fact that we do not want the GP to rely on solutions that previously converged on local optima.

\begin{figure*}[ht!]
\centering
\includegraphics[width=.9\textwidth]{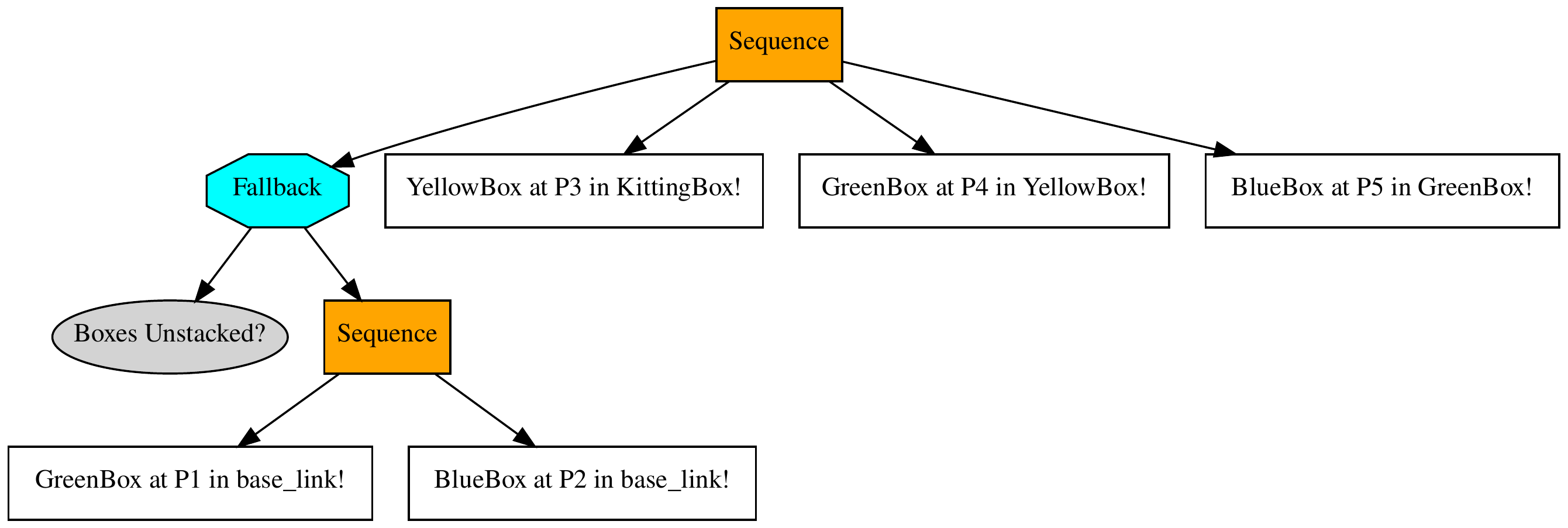}
\caption{BT solving the task defined in Experiment~3. First the boxes are unstacked and placed somewhere in the base frame of the robot. Then, they are stacked in the target configuration at the center of the \textit{KittingBox}, as shown in Figure~\ref{fig:exp1}. The white boxes represent a subtree in the form of Figure~\ref{fig:gene}, where $P_{1,2}$ are the position chosen by the user in demonstration and $P_{3,4,5}$ are the position of the boxes in the configuration of Figure~\ref{fig:exp1}.}
\label{fig:exp3_solution}
\end{figure*}

With the demonstrated BT, the GP is able to successfully solve the task (Figure~\ref{fig:res3}). A compacted version of the BT solving the task, where we collapsed a subtree responsible for picking and placing a box in a single behavior, is reported in Figure~\ref{fig:exp3_solution}.

\begin{figure*}[ht!]
\centering
\begin{subfigure}{0.34\textwidth}
\includegraphics[width=\textwidth]{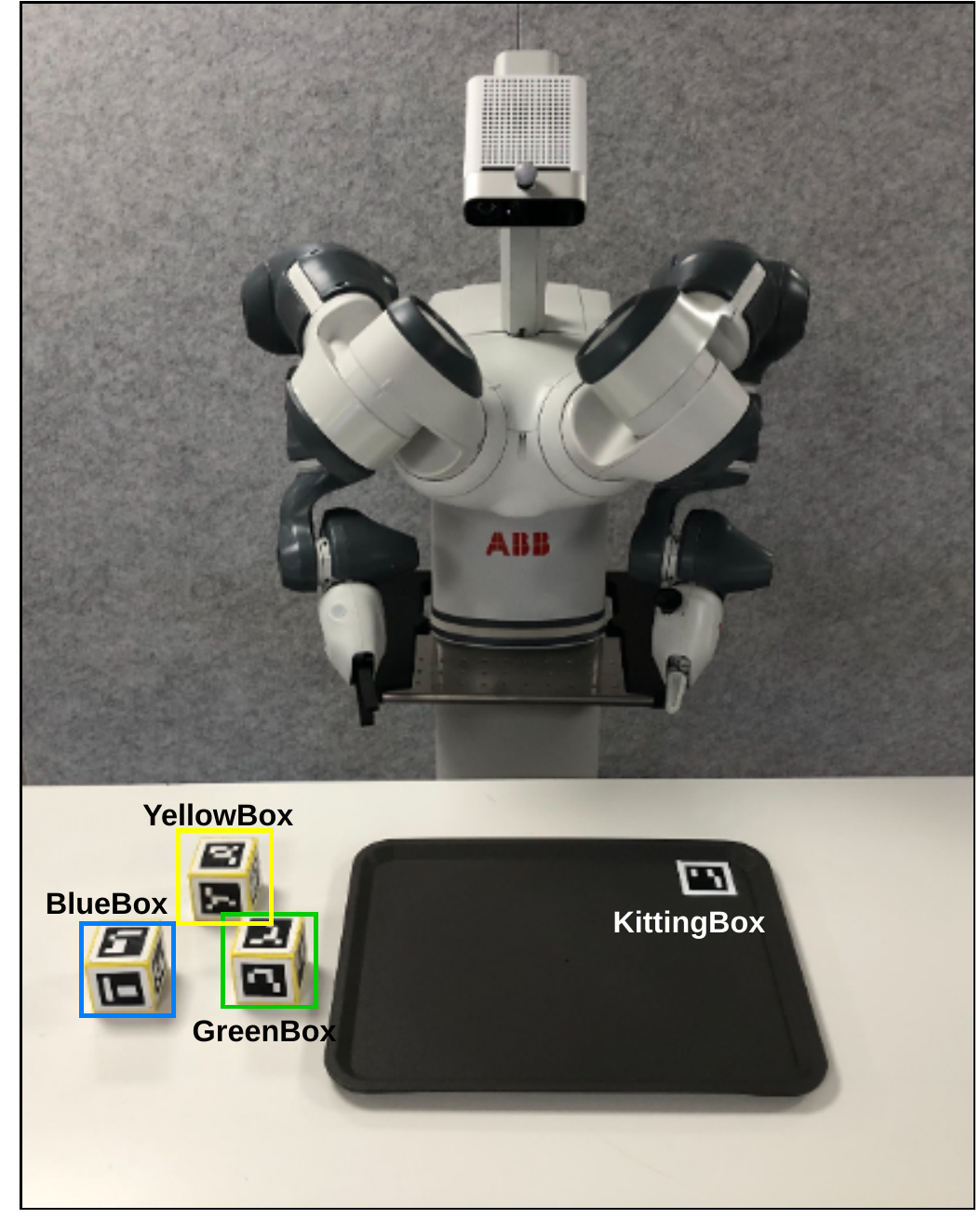}
\caption{Initial configuration of the task.}
\label{fig:real1}
\end{subfigure}
\begin{subfigure}{0.28\textwidth}
\includegraphics[width=\textwidth]{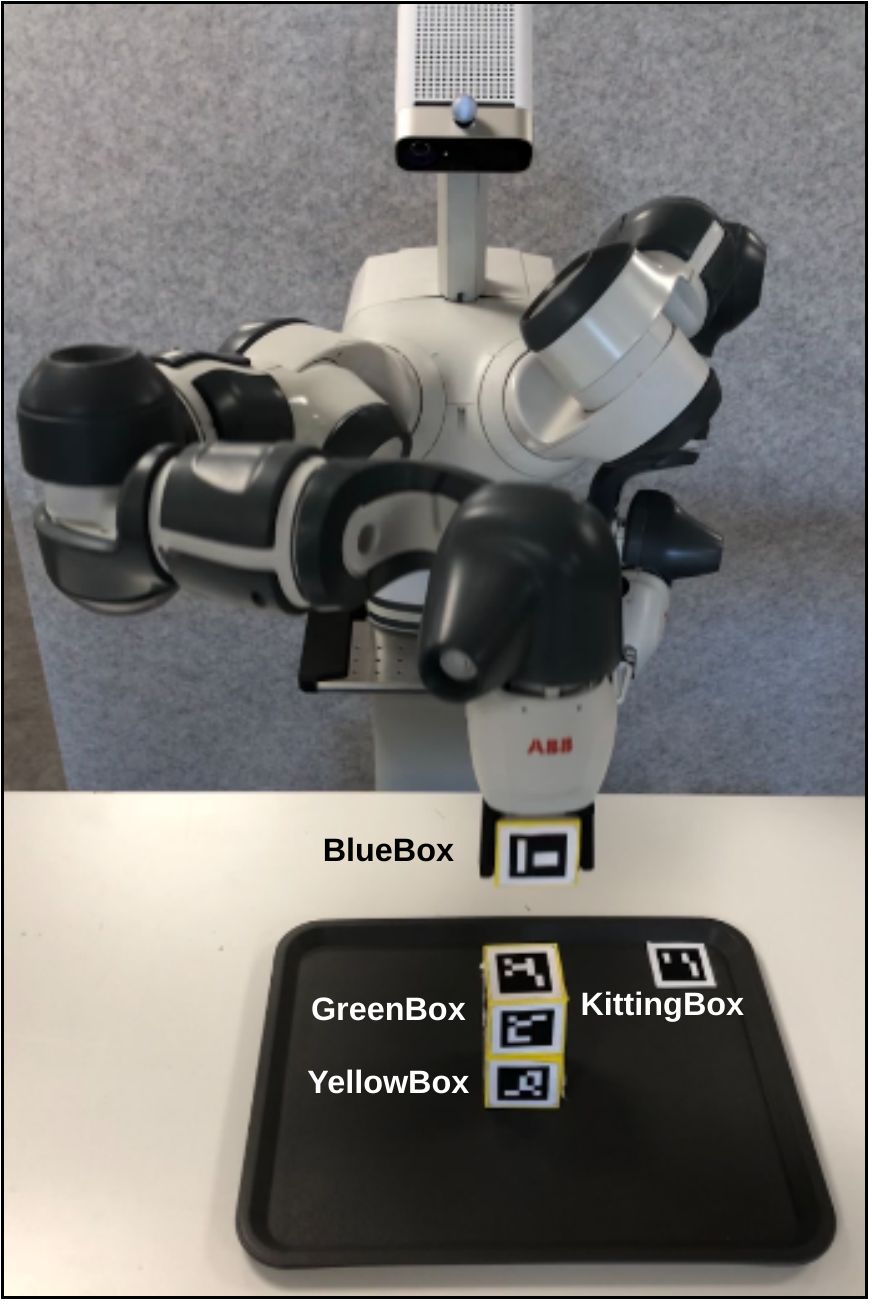}
\caption{The robot places the \textit{BlueBox}.}
\label{fig:real2}
\end{subfigure}
\begin{subfigure}{0.33\textwidth}
\includegraphics[width=\textwidth]{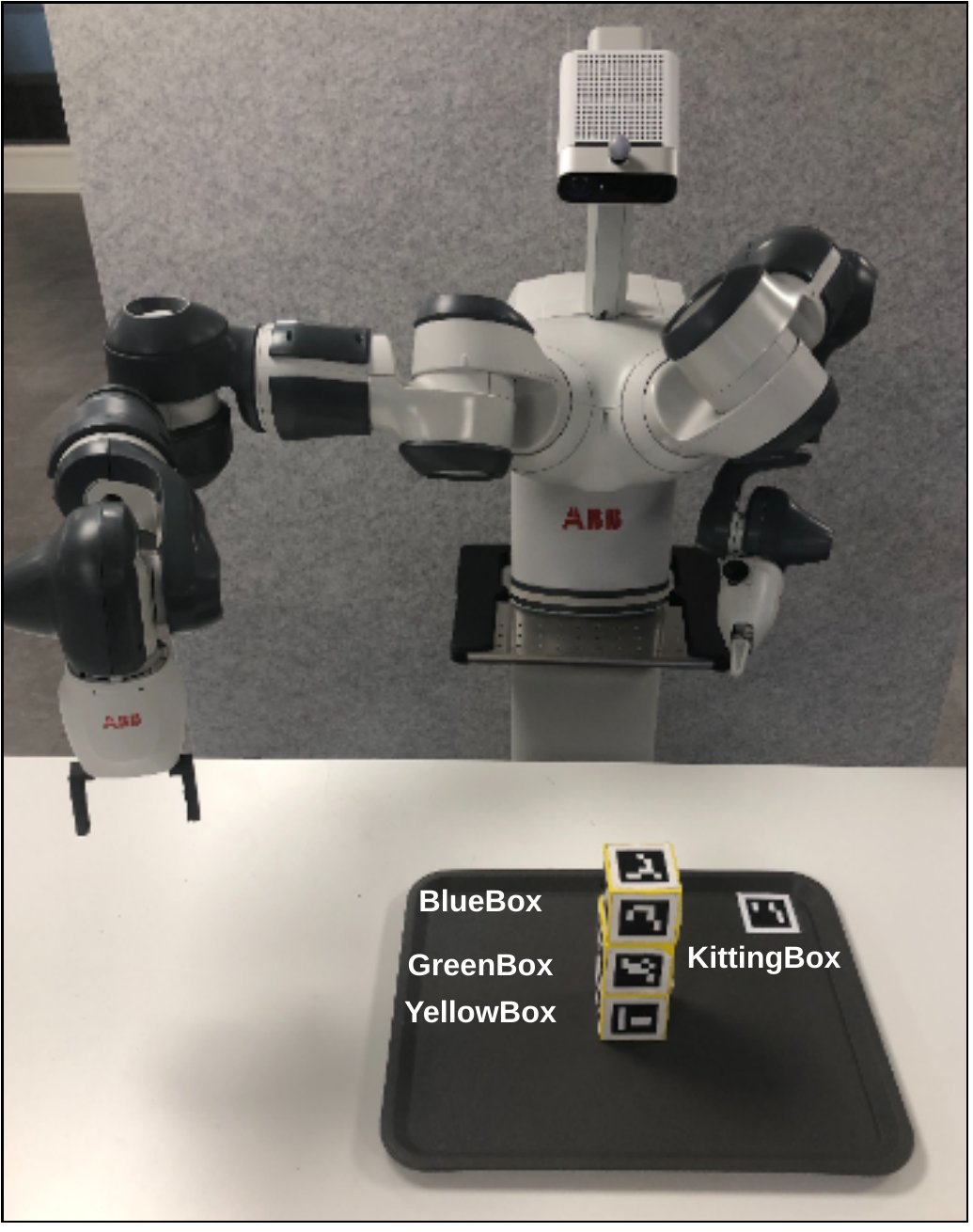}
\caption{Final configuration of the task.}
\label{fig:real3}
\end{subfigure}
\caption{In Experiment~4 we transfer the solution of Experiment~1 on the real robot.}
\label{fig:real}
\end{figure*}

\paragraph{\textit{Experiment 4}} \label{exp4}

In this experiment we transfer the solution of Experiment~1 on a real robot. The robot is an ABB YuMi with an Azure Kinect camera. For the perception we use Aruco markers detection. Every box has a different marker in any of the faces, so that we are able to generate a reference frame for each box in its centroid. Another marker is attached to the \textit{KittingBox}. In this case, we place it in a corner to avoid occlusion but we generate a reference frame at its center using offsets along the $X$ and $Y$ axis. 

\textbf{Results:} The BT learnt in simulation can be used without modification on a real scenario to successfully solve the same task (Figure~\ref{fig:real}). By using a BT to control the robot, we make it reactive to changes in the environment. If an operator helps the robot in building the stack or undoes parts of it, the robot reacts by either completing the remaining steps of the tasks or redoing it, as demonstrated in more detail in~\cite{gustavsson_combining_2022}.

\section{Conclusions and Future Work} \label{sec:conclusion}

%\begin{itemize}
%    \item allow the possibility to input BTs learnt by demonstrating a task on the real robot
%    \item extend to mobile manipulation tasks
%    \item extend to different level of fidelity for the simulator
%\end{itemize}

In this paper we proposed to combine a method that evolves Behavior Trees (BTs) using Genetic Programming (GP) that we proposed in~\cite{iovino_learning_2021}, with a method that learns BTs from demonstration that we proposed in~\cite{gustavsson_combining_2022}. We showed that the combined framework allows users with low to no programming skills to automatically, intuitively, and time-efficiently generate BTs for robotic applications. The method learns BTs in an unsupervised fashion but can exploit human experience in task solving in the form of demonstrations. We illustrated the strengths of this method in a series of simulated manipulation tasks. We transferred one of the learnt solutions to a real robot, showing that we can learn in simulation without losing generality. In this paper we targeted collaborative robotic applications but the framework can be extended to other robotic tasks, for instance industrial robotic ones, since the demonstrations and the learning process are performed in simulation.

As future work, we plan on applying the learning framework to mobile manipulation tasks. Moreover, we will exploit the design freedom of the simulator to define a higher level of fidelity. In this way, we propose to model faults to force the framework to learn more robust solutions. As an example, we can simulate a simple model of a mobile platform and constrain the gripper to a fixed workspace. As a consequence, if an item to pick is not reachable, the robot will have to approach it first.

This framework is meant to allow non-expert programmers to generate BTs intuitively and time-efficiently. Therefore, once we have further developed the framework to tackle a wide range of mobile manipulation tasks, we aim to test its usability with an user study. We expect that the outcome of this study will give some insights on the design of the graphical user interface and guide its development.

\section*{Acknowledgments}
This work was carried out in the WASP Research Arena (WARA) - Robotics, hosted by ABB Corporate Research Center in Västerås, Sweden and financially supported by the the Wallenberg AI, Autonomous Systems, and Software Program (WASP) funded by the Knut and Alice Wallenberg Foundation. The authors thank Daniel Lindmark from Algoryx for the valuable support provided with the simulator.

\bibliographystyle{ieeetran}
\bibliography{references.bib}

\end{document}